\documentclass[sn-nature]{sn-jnl}

\usepackage{graphicx}%
\usepackage{multirow}%
\usepackage{amsmath,amssymb,amsfonts}%
\usepackage{amsthm}%
\usepackage{mathrsfs}%
\usepackage[title]{appendix}%
\usepackage{xcolor}%
\usepackage{textcomp}%
\usepackage{manyfoot}%
\usepackage{booktabs}%
\usepackage{algorithm}%
\usepackage{algorithmicx}%
\usepackage{algpseudocode}%
\usepackage{listings}%
\usepackage{orcidlink}
\usepackage{subfigure}
\usepackage{natbib}
\usepackage{hyperref}

\theoremstyle{thmstyleone}%
\theoremstyle{thmstyletwo}%

\theoremstyle{thmstylethree}%

\begin{document}

\title[Article Title]{Automated and Interpretable Survival Analysis from Multimodal Data}

\author*[1]{\fnm{Mafalda} \sur{Malafaia}\orcidlink{0000-0002-8081-0454}}\email{Mafalda.Malafaia@cwi.nl}

\author[1,2]{\fnm{Peter A. N.} \sur{Bosman}\orcidlink{0000-0002-4186-6666}}

\author[3]{\fnm{Coen} \sur{Rasch}\orcidlink{0000-0001-6950-3376}}

\author[3]{\fnm{Tanja} \sur{Alderliesten}\orcidlink{0000-0003-4261-7511}}

\affil[1]{\orgname{Centrum Wiskunde \& Informatica}, \orgaddress{\city{Amsterdam}, \country{The Netherlands}}}
\affil[2]{\orgname{Delft University of Technology}, \orgaddress{\city{Delft}, \country{The Netherlands}}}
\affil[3]{\orgname{Leiden University Medical Center}, \orgaddress{\city{Leiden}, \country{The Netherlands}}}

\abstract{

Accurate and interpretable survival analysis remains a core challenge in oncology. With growing multimodal data and the clinical need for transparent models to support validation and trust, this challenge increases in complexity. We propose an interpretable multimodal AI framework to automate survival analysis by integrating clinical variables and computed tomography imaging. Our MultiFIX-based framework uses deep learning to infer survival-relevant features that are further explained: imaging features are interpreted via Grad-CAM, while clinical variables are modeled as symbolic expressions through genetic programming. Risk estimation employs a transparent Cox regression, enabling stratification into groups with distinct survival outcomes. Using the open-source RADCURE dataset for head and neck cancer, MultiFIX achieves a C-index of 0.838 (prediction) and 0.826 (stratification), outperforming the clinical and academic baseline approaches and aligning with known prognostic markers. These results highlight the promise of interpretable multimodal AI for precision oncology with MultiFIX.
}


\maketitle

\section{Introduction}\label{introduction}

Head and neck cancer (HNC) is the sixth most common type of cancer worldwide, with incidence rates continuing to rise globally and over 16,000 deaths in 2024~\cite{nguyen2025metastatic}. Known prognostic factors include age, tumor stage, smoking, co-morbidity, weight loss, and performance status~\cite{leoncini2015clinical}. However, there is a need for effective prognostic tools to accurately guide clinical decision-making and optimize treatment planning.

The currently accepted clinical guideline for HNC prognosis relies on the TNM staging system, which categorizes tumors according to tumor size~(T), lymph node involvement~(N), and distant metastasis~(M), to stratify patients into different risk groups~\cite{edge2010american}. While this stratification technique correlates with different prognoses per group and is successfully employed in clinical practice~\cite{mesia2010seom}, there are some limitations that may hinder its accuracy: the TNM staging system does not factor in the inherent heterogeneity of HNC patients, providing limited individualized prognostic information; furthermore, it fails to account for other important prognostic factors that may influence the progression of the disease~\cite{tnm_usage, chow2020head, tham2019personalized}. This is particularly emphasized when promoting precision medicine, in which treatment decisions increasingly require personalized risk assessment according to patient-specific characteristics beyond TNM staging. Clinicians currently address these challenges using other known prognostic factors and their past experience to complement the prognostic information given by the TNM survival curves.

In addition to the TNM staging systems for survival analysis, Cox regression models are a standard approach to evaluate prognostic factors and predict the risk of event occurrence~\cite{cox_literature}. However, in these models, a linear relationship between the survival estimates and the input clinical variables is assumed, which may impose limits on the ability to make accurate predictions. Furthermore, these models do not provide meaningful insights into how to stratify patients into representative subgroups; instead, they focus on whether a specific variable has a positive (protective) or negative (harmful) effect on prognosis. Considering the properties of this approach, Cox regression models are highly valuable in clinical trials to determine whether a certain variable increases the likelihood of survival over time~\cite{hajdu2022swallowing, runge2023perioperative, clarke2020head}, rather than focusing on building high-performing predictive models.

The aforementioned limitations of traditional prognostic approaches have driven the development of artificial intelligence~(AI) models to perform survival analysis for several types of cancer. Recent literature demonstrates that deep learning~(DL) models consistently outperform traditional Cox regression models by overcoming the linearity constraint imposed by the latter. Notable mentions include DeepSurv~\cite{deepsurv}, a DL approach that optimizes the model by minimizing the negative logarithm of the partial likelihood from the Cox model. Furthermore, recent studies increasingly report the integration of multimodal data into survival analysis pipelines in order to exploit complementary information from different data sources~\cite{dl_surv}. Although multimodal DL approaches show promising gains in predictive accuracy, the clinical translation of AI-based survival analysis models remains limited due to a critical barrier: their lack of interpretability.
Healthcare professionals seek not only accurate predictions but also understandable explanations to foster trust and enable verifiability in computer-assisted decision-making~\cite{mohamed2025decoding}. This need for interpretability is particularly critical in survival analysis, where treatment decisions based on prognostic assessments can have profound implications for patient outcomes and quality of life~\cite{langbein2024interpretable}.

Despite most survival analysis approaches lacking interpretability, there have been some efforts to develop interpretable models. Traditional approaches have focused on adapting post-hoc explanation methods to survival contexts, such as  SurvSHAP~\cite{alabdallah2022survshap}, which extends SHAP (SHapley Additive exPlanations) to survival outcomes, and SurvLIME~\cite{kovalev2020survlime}, which adapts Local Interpretable Model-agnostic Explanations for time-to-event data. However, these often require surrogate models to handle the complex survival curve outputs.
In the multimodal domain, interpretability efforts have shown promise but remain limited in scope. Notable mentions of interpretable multimodal approaches for survival analysis include a multi-task logistic regression~(MTLR)~\cite{baseline-paper}, which is used to predict overall survival for HNC patients using computed tomography~(CT) scans and clinical information as input. This approach claims interpretability in the models with time-specific Grad-CAM heatmaps that highlight relevant regions in the images for each risk prediction. Unfortunately, the influence of the clinical variables is disregarded in the explanations, as is the influence of each modality on risk prediction. Additionally, two risk groups are established using the median risk as a threshold; however, the performance of these groups is not calculated, and the group stratification is not interpretable, demonstrating an overall limited use for clinical purposes regarding patient risk stratification. Another interpretability-focused approach was developed in~\cite{eijpe2025disentangled}, where a multimodal transformer-based pipeline is used to predict disease-specific free survival using whole slide images~(WSI) and transcriptomics data as input. The pipeline includes a feature disentanglement module that is used to separate features shared between modalities and modality-specific features. Each data modality is provided with a unimodal feature extraction step that includes modality-specific interpretability techniques. 
Despite the efforts to extract explainable features per modality, the approach lacks focus on risk prediction interpretability, as SHAP~\cite{shap2017} is used exclusively to quantify the contributions of modality-specific and modality-shared features. However, there is no further analysis regarding clinical relevance. In the mentioned work, patient risk stratification is not studied.

Risk stratification is not only underexplored in AI-based survival but is also directly tied to clinical decision-making, as treatment planning in oncology often depends on stratifying patients into distinct risk groups. DL approaches tend to either focus on risk prediction or use the median risk output to create subgroups. Survival trees~\cite{survival1992trees}, on the other hand, are a powerful and interpretable approach to clustering patients into different groups according to survival outcomes by recursively partitioning the data based on the input variables while handling censored data. The integration of survival tree optimization using DL architectures showed great potential for risk stratification~\cite{sun2024survrelu}. Despite their potential, survival trees are limited in capturing non-linear relationships when used independently, highlighting the need to integrate feature engineering with tree optimization. 

This work addresses the critical gap between AI model performance and clinical interpretability in survival analysis by using MultiFIX, a Multimodal Feature engIneering eXplainable AI framework specifically designed to provide insightful models that can be useful in the clinical setting. 
Our MultiFIX-based framework for survival analysis makes three main contributions:
\begin{enumerate}
    \item Interpretable feature engineering per data modality using DL and modality-specific interpretability methods;
    \item Feature importance assessment and interpretable risk prediction with the use of previously engineered features as input in a Cox regression model;
    \item Interpretable patient risk stratification to generate representative survival curves that can be assigned to each patient according to the engineered features.
\end{enumerate}

MultiFIX leverages the integration of DL and genetic programming~(GP) to provide accurate yet interpretable models. While DL is used to train the full model and simultaneously engineer relevant features from each data modality, GP is used to replace the tabular feature engineering block with an inherently interpretable symbolic expression. Patient risk stratification is introduced in the pipeline to tackle the main research gap in survival analysis, as current literature shows limited work on this topic. The key contributions of this work include: (1) the development of a novel interpretable multimodal framework for survival prediction; (2) the use of GP to generate interpretable symbolic expressions for clinical feature engineering; and (3) the demonstration of the generated interpretable model for HNC overall survival analysis based on both clinical variables and CT imaging, along with a subsequent assessment of the clinical relevance of the model.

By addressing the interpretability challenge while maintaining competitive predictive performance, this work represents a significant step towards the clinical translation of AI-powered survival analysis. Our MultiFIX-based pipeline demonstrates the feasibility of deploying transparent AI in oncology workflows by integrating advanced analytical techniques with clinical interpretability requirements. The head and neck cancer use case serves to illustrate the potential application of this framework in real-world prognostic assessment and treatment planning, rather than representing a focused innovation for this specific disease context.

\section{Results}\label{results}

This section presents the results we obtained through models trained with our MultiFIX-based pipeline. These models were trained using two data configurations that we refer to as core and extended data configurations. The former is comparable to that used in~\cite{baseline-paper}, whose results we adopt as our comparison baseline. The latter extends the core dataset with clinical variables and a downsampled version of the CT images as an additional modality. Both configurations included a total of 1,653 patients for training. An external test cohort of 670 patients was used to evaluate the resulting models. Further information on these datasets is available in Section~\ref{sec:methods}.

This section presents the findings from the models learned using our MultiFIX-based pipeline. The results are organized into the following subsections: Subsection~\ref{subsec:features} details the features that the learned models engineered for each of the data modalities; Subsection~\ref{subsec:performance} presents the performance results of the learned models and the corresponding baseline performances for both risk prediction and stratification; Subsection~\ref{subsec:interpretability} provides an interpretability analysis of the resulting interpretable model, specifically the explainable engineered features, the survival analysis achieved by the Cox regression model, and the resulting patient stratification into risk groups; lastly, Subsection~\ref{subsec:clinical} presents an analysis of the clinical findings inferred from the model with expert consultation.

\subsection{Features Engineered by the learned models}
\label{subsec:features}

This subsection describes the features engineered by the best-performing model and how these are interpreted. Image-engineered features are extracted by the DL image feature engineering block and explained using activation maps generated with Grad-CAM. The tabular feature engineering block is used to extract the engineered features and is then replaced with symbolic expressions that approximate the DL features. The number of engineered features per feature engineering block is selected through hyperparameter optimization~(HPO), as further described in Section~\ref{sec:methods}

\subsubsection{Features engineered from CT imaging}

Two different 3D CT imaging inputs were used to engineer features from: (1) the tumor imaging input, where the original CT is resampled to a pixel spacing of $1~mm$, centered in the center of mass of the Gross Tumor Volume~(GTV), and cropped to $128\times128\times64$ voxels; (2) the downsampled imaging input, where the original CT is resampled to a pixel spacing of $3~mm$ and cropped using its own center of mass to $128\times128\times64$ voxels. The architecture of the learned models included one feature engineering block for each of these imaging inputs: the tumor image feature engineering block and the downsampled image feature engineering block.

For the core data configuration, the MultiFIX model with the highest Concordance index~(C-index) in the validation set presented a bottleneck of 3 engineered features~(denoted by $Ti$) for the tumor image feature engineering block, each presenting the corresponding value range: $T1$ is in the range $[0.365, 0.718]$; $T2$ is in the range $[0.352, 0.724]$; $T3$ is in the range $[0.444, 0.606]$. The respective activation maps for each feature show high variability per sample and feature. Feature $T2$ heatmaps show consistently high activations in the GTV region and its surroundings, while features $T1$ and $T3$ appear to have higher contributions in other potentially relevant structures in the image. Images and corresponding features and activation maps are available in the Figure \ S4 (see Supplementary Material).

For the extended data configuration, the best model has a bottleneck of 3 features for the downsampled images, with the following feature ranges: $[0.264, 0.531]$ for $D1$; $[0.183, 0.447]$ for $D2$; $[0.209, 0.641]$ for $D3$. Optimizing the number of features per modality revealed that the prediction model performed best when excluding the tumor images from the input. This can be related to the redundancy stemming from having two imaging feature engineering blocks pertaining to the same modality (CT scan), since the downsampled image, although having a lower resolution, can include information from the tumor image. The activation maps from each of the engineered features derived from the downsampled images revealed important contributions stemming from several areas of the image that are not necessarily related to the tumor but predominantly include soft tissue and muscle tissue. When used by a clinician, the activation maps from different samples and their respective CT scans would be provided so that they could further confirm whether the most activated regions of the images were relevant to the risk assessment.

\subsubsection{Features engineered from clinical variables}

The model engineered tabular features from 10 clinical variables for the core data configuration and from 16 variables for the extended data configuration.

\emph{Core Data Configuration:} The final model learned from the core data configuration comprised a bottleneck of 2 clinical features~($C1$ and $C2$), with a feature range of $[0.000, 1.000]$ for both.

GP-GOMEA~\cite{virgolin2018symbolic} was used to generate a total of 15 symbolic expressions, with 5 for each of the studied complexities (expressed as maximum tree depth). The expressions show several similarities in terms of the clinical variables used. For feature $C1$, age and smoking status are the most prevalent variables, appearing in 14 and 13 expressions, respectively. Other frequent variables include human papillomavirus~(HPV) status, concurrent chemoradiotherapy~($chemo$), TNM stage~($stage$), and Eastern Cooperative Oncology Group Performance Status (ECOG PS). The sex of the patient was not used in any of the identified expressions. For feature $C2$, TNM stage~($stage$) was the most frequent variable, appearing in 14 expressions and being exclusively used in 7 of those expressions. Other variables were used only in some expressions, namely smoking status and the tumor size (T). The best equation for each maximum tree depth was selected according to the lowest Mean Squared Error~(MSE) on the test set, in comparison with the DL-engineered features, translating to the expressions presented in Table~\ref{tab:gp_expressions_baseline}.

The resulting symbolic expressions show a consistent decrease in MSE as the maximum tree depth increases, indicating that the more complex the symbolic expression, the more similar the GP feature is to the DL feature. Despite its increased complexity, at depth 4, the evolved symbolic expressions are still readable and provide valuable insights into how each clinical variable contributes to the feature value. We will focus our analysis on the features with a complexity of depth 4, further denoted as $C1_{d4}$ and $C2_{d4}$. Further analysis of how these features behave according to different values of the clinical variables used is provided in Figures \ S1, S2, and S3 (see Supplementary Material).

Feature $C1_{d4}$ models the joint effect of several clinical variables using two distinct terms: the first term correlates a higher feature value with lower smoking status (non-smoker:$1$; ex-smoker:$0$; current smoker:$-1$) and HPV status (positive:$1$; unknown:$0$; negative:$-1$) values; the second term increases the feature value for patients receiving chemoradiation treatment ($chemo = 1$), higher $ECOG_{PS}$ values (with values ${0,1,2,3}$, where $0$ indicates fully active and $3$ limited to no self-care ability), and older patients (higher $age_{norm}$). The overall behavior of $C1_{d4}$ is consistent with clinical knowledge: smoking and age are strong prognostic factors in HNC; HPV-positive status is usually correlated with a better prognosis; ECOG PS is correlated with a worse prognosis; chemoradiation, although not directly correlated with a worse prognosis, is often chosen for more aggressive tumors, reflecting that patients who receive treatment usually have worse prognoses.

Feature $C2_{d4}$ models the relationship between the TNM Stage and other variables: the TNM Stage influences the feature value the most, being directly correlated with $C2_{d4}$; age, smoking status, HPV status, and chemoradiation treatment are directly correlated with a lower feature value. A constant is included in the denominator to stabilize the feature value and decrease the influence of these variables. While the overall behavior of $C2_{d4}$ is correlated with clinical knowledge indicating a worse prognosis, the age and chemoradiation variables seem to present conflicting effects on the feature value. It can be hypothesized that the influence of these variables on the feature is modulated by the constant present in the denominator and is even overpowered by the Stage variable.
 
\emph{Extended data configuration:} The final model learned from the extended data configuration comprised a bottleneck of a single clinically-engineered feature, with a value range of $[0.237, 0.806]$.

Similarly to the core data configuration, GP-GOMEA generated 15 symbolic expressions with different complexities. The expressions show not only higher MSE values for all complexities but also greater disparities in the variables used. Smoking status and the T stage were the most commonly used variables, followed by age and chemoradiation treatment. Other used variables include the number of delivered radiotherapy~(RT) fractions and the ECOG PS. The best solutions per maximum tree depth are presented in Table~\ref{tab:gp_expressions_extended}.

Contrary to the core data configuration, there is no clear decrease in MSE as tree depth increases, namely from depth 3 to depth 4, since both symbolic expressions achieve similar performance using the DL feature as a reference. The extended configuration model used depth 3 expressions. The selected feature, $C1_{d3}$, shows a direct correlation with the T stage, the age of diagnosis (normalized between $0$ and $1$), and the TNM stage, and an inverse correlation with chemoradiation treatment and smoking status. The numerator of the present fraction shows a relationship between $T$ and $chemo$, where the administration of chemoradiation treatment suggests an improvement in the tumor stage. This correlates with clinical knowledge, as the treatment is administered to patients in more advanced stages (mainly T3 and T4). Age is a known prognostic factor for overall survival, as is smoking status. Finally, the more advanced the TNM Stage, the lower the denominator of the fraction, leading to higher feature values. Overall, feature $C1_{d3}$ is strongly correlated with worse prognostic factors.

\subsection{Performance Results}
\label{subsec:performance}

This subsection is organized according to the two different tasks assessed by our survival analysis pipeline. The interpretable models learned with our MultiFIX-based pipeline are used to make individual risk predictions for each patient (Risk Prediction). The features engineered by the model, along with the corresponding predictions, are then used to perform patient stratification into distinct risk groups (Risk Stratification). Both tasks are evaluated in terms of the average C-index and the area under the receiver operating characteristic curve~(AUROC) at a 2-year follow-up~(AUC2yr), with the respective 95\% confidence intervals computed using 1000 sample bootstrapping.

\subsubsection{Risk Prediction}

Table~\ref{tab:risk_performance} summarizes the performance of various risk prediction models evaluated using the C-index and AUC2yr metrics, with 95\% confidence intervals.

Baseline results are included in the table for comparison, namely for the use of single modality DL models for both the tumor and downsampled imaging inputs, as well as the use of a Cox regression model with the clinical variables as input. The results of the multimodal multitask logistic regression approach~(MM MTLR) are also indicated using the performance results from the publication~\cite{baseline-paper}. These results show consistent multimodal improvements compared to all data modalities. All interpretable models (using Cox regression and GP clinically-engineered features) show great predictive potential, achieving competitive performance compared to the original DL model and the baseline reference (MM MTLR). The performance-interpretability trade-off seems marginal for models with symbolic expressions of tree depths 3 and 4.

Additionally, a Cox regression model was fitted using both the combination of the engineered DL features and the original clinical variables. The performance results matched those of the MultiFIX DL model, suggesting that the Cox regression model uses the engineered features for its predictions.

Despite the additional information, the models learned using the extended data configuration are consistently outperformed by the models learned with the core data configuration once the survival (fusion) block and the tabular feature engineering block are replaced by their interpretable counterparts. Performance results from the DL version of the models show similar average performance for both approaches. This may indicate that the model learned from the extended data configuration may be using more complex interactions between engineered features to make the final prediction, thus achieving lower results once the DL survival block is replaced with a Cox regression, which has linearity constraints. Still, the performance, both in terms of C-index and AUC2yr shows very small variance between the two data configurations, and the confidence intervals have very similar ranges. Thus, one could argue that the models have very similar performance, suggesting that the information added to the extended data configuration does not improve the predictive ability of the model.

\subsubsection{Risk Stratification}

Patient stratification was employed using quantile stratification based on the risk predictions from each MultiFIX model to create groups of patients. Table~\ref{tab:group_performance} refers to the performance results pertaining to the stratified versions of several MultiFIX models in terms of C-index and AUC2yr, along with the respective 95\% CI. TNM staging refers to the clinical baseline currently used in clinical practice, using 6 groups, one for each of the following stages: $0,~I,~II,~III,~IVA,~\text{and }IVB$. The number of groups chosen for each model risk stratification relied on the pairwise statistical significance between survival curves from each group using log-rank testing, up to a maximum of 6 groups to match the clinical baseline.

For the TNM staging stratification, only groups $IVA$ and $IVB$ are significantly separable; the remaining consecutive pairs of groups show a p-value lower than $0.05$ in the log-rank test. For all MultiFIX models, stratification resulted in 6 significantly separable risk groups.

Performance-wise, all MultiFIX models show substantial improvements in C-index compared to TNM Staging Stratification. For the MultiFIX model using Cox regression with DL-engineered image features and GP-engineered tabular features of depth 4, performance improved by $0.212$ in C-index for the core data configuration. With the extended data configuration, the MultiFIX model using Cox regression with DL-engineered image features and GP-engineered tabular features of depth 3 shows similar, though smaller, improvements, yielding a C-index of $0.787~(0.750, 0.825)$.

The quantile stratification approach was compared with the use of survival trees built on the engineered features to create patient groups. Nonetheless, the results showed that the former approach achieves consistently higher C-index values and greater separability between a larger number of risk groups. Performance results for survival tree stratification are presented in Table \ S3 (see Supplementary Material).

\subsection{Interpretability Analysis Results}
\label{subsec:interpretability}

To showcase the interpretability of the resulting models, we selected the MultiFIX model with GP-engineered features of depth 4, as this model was able to match DL performance while still remaining readable for the user. In a clinical setting, these models would be presented to clinicians so that they could opt for the model of their preference.

For the core data configuration, the model included three features engineered from the tumor images~(denoted as $Ti$ in the model) and two features engineered from the clinical variables (denoted as $Ci$ in the model). The final interpretable model is presented in Figure~\ref{fig:xai_model_baseline}.

For the extended data configuration, the model included three features engineered from the downsampled image modality~(denoted as $Di$ in the model) and one feature engineered from the clinical variables (denoted as $C1$ in the model). The interpretability analysis section focuses on the core data configuration model since, despite including less information, it achieves competitive performance and uses a 3D subvolume of the CT scan cropped around the GTV with a higher resolution than the downsampled image inputs (pixel spacing of $1~mm\times1~mm\times1~mm$ for tumor images and $3~mm\times3~mm\times3~mm$ for downsampled images), providing high-resolution information at the primary tumor region. For completeness, the interpretable model for the extended data configuration is presented in Figure \ S5 (see Supplementary Material).

\subsubsection{Risk Prediction}

The actual survival risk prediction occurs in the Survival Fusion Block. An illustration thereof is provided in  Figure~\ref{fig:xai_model_baseline}~3A. The Cox regression model output is presented in a table containing, for each engineered feature, the hazard ratio~(HR), followed by its confidence interval and a p-value that indicates whether the feature has a statistically significant effect on the prediction. The features are color-coded in green if the HR indicates a protective effect and in red if the HR indicates a harmful effect.

For the core data configuration, the Cox regression model indicates that two of the MultiFIX-engineered features are protective, and three are harmful. Feature $T2$ seems to have the strongest negative impact on the risk of the event (death), followed by $C1$ and $C2$. $T1$ and $T3$ seem to have a similar positive impact on the event, indicating that the higher their values, the lower the risk prediction.

For the extended data configuration (Figure \ S5, see Supplementary Material), the Cox regression model indicates one protective feature ($D3$ with $HR = 0.56$), two harmful features ($D1$ with $HR = 1.43$ and $C1$ with $HR = 1.55$), and a feature that did not show a significant association with overall survival (p-value$>0.05$). 

\subsubsection{Risk Stratification}

\begin{figure}
    \centering
    \includegraphics[width=0.99\linewidth]{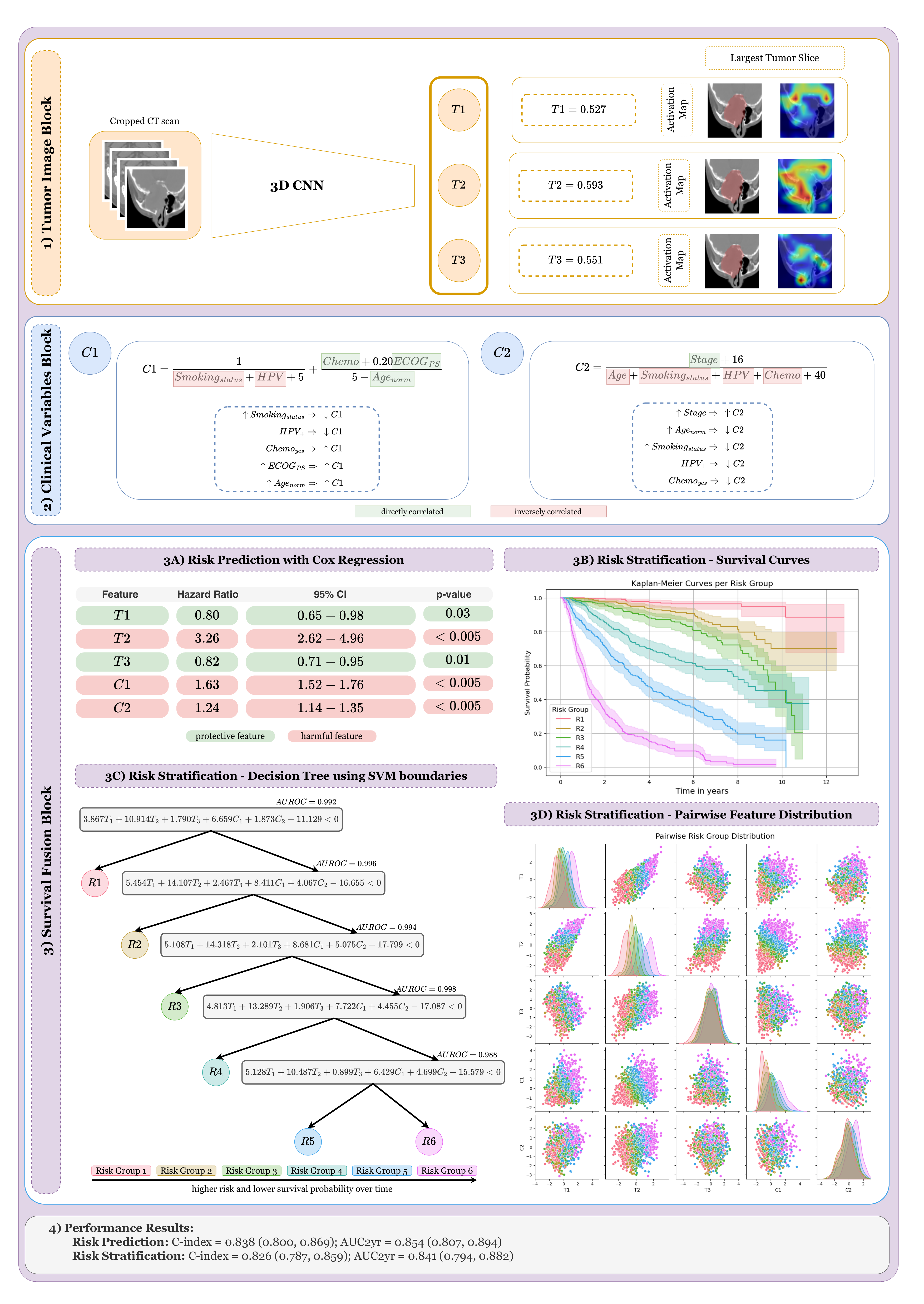}
    \caption{Interpretable multimodal survival analysis model using the MultiFIX-based framework with the core data configuration. (1) Tumor Image Block: 3D convolutional neural network feature engineering block using cropped CT scans as input to extract interpretable tumor features ($T1$, $T2$, $T3$) with corresponding Grad-CAM activation maps highlighting the largest tumor slice (as an example). (2) Clinical Variables Block: two GP-GOMEA generated symbolic expressions represent the engineered features with the original clinical features: C1 incorporates smoking status~($smoking_{status}$), HPV status~($HPV$), chemoradiation~($chemo$), ECOG PS~($ECOG_{PS}$), and normalized age~($age_{norm}$); C2 combines TNM stage ($stage$), normalized age~($age_{norm}$), smoking status~($smoking_{status}$), HPV status~($HPV$), and chemoradiation~($chemo$). Directional arrows and color-coding indicate positive ($\uparrow$ or green) or inverse ($\downarrow$ or red) correlations between individual variables and composite features. (3) Survival Fusion Block: Cox proportional hazards regression (3A) quantifies feature contributions through hazard ratios, distinguishing protective (HR $<$ 1.0) from harmful (HR $>$ 1.0) features. Risk stratification results are used to generate Kaplan-Meier survival curves (3B), decision tree boundaries using SVM boundaries between groups (3C), and pairwise risk group distributions (3D) to demonstrate model interpretability across six risk groups (R1-R6). The risk group color legend is consistent across subfigures 3B, 3C, and 3D. Model performance is indicated in (4).}
    \label{fig:xai_model_baseline}
\end{figure}

The risk stratification assessment is performed with three different considerations: in Figure~\ref{fig:xai_model_baseline}~3B, the survival curves for each risk group are plotted in combination with their confidence intervals; in Figure~\ref{fig:xai_model_baseline}~3D, a pairwise plot is presented to illustrate the group distribution according to the engineered features; in Figure~\ref{fig:xai_model_baseline}~3C, a decision tree is presented using support vector machines~(SVM) boundary equations to assign patients to different risk groups.

For the core data configuration, the Kaplan-Meier survival curves show different survival behaviors for various risk groups, with mostly good separation. As expected, groups 5 and 6 show the poorest survival, while groups 1 and 2 exhibit the lowest mortality rates. The pairwise feature distributions~((Figure~\ref{fig:xai_model_baseline}~3D) over groups show different behaviors for different pairs of features. The diagonal presents the distributions of the groups according to each individual feature. Clearly, no single feature is sufficient to properly stratify the patients. Feature $T2$ reveals the highest separability between groups. This was expected since it presents the highest HR. Several interactions between features correlate with the risk distribution. This is particularly the case for the combination of $T2$ and $C1$, for which well-separated groups are identified, separated along diagonals that indicate the interaction between the features. Lastly, the decision tree allows for assigning a patient to a risk group according to the extracted feature values in a transparent manner. The decision nodes are equivalent to the boundaries learned with an SVM classifier between each pair of consecutive risk groups and allow for easy separation between groups, considering the feature values that can be extracted with the MultiFIX model.

The MultiFIX model learned with the extended data configuration showed similar results and is presented in Figure \ S5 (see Supplementary Material).

\subsection{Clinical relevance of the learned models}
\label{subsec:clinical}

\begin{figure}
    \centering
    \subfigure[TNM Staging Stratification]{
        \includegraphics[width=0.44\textwidth]{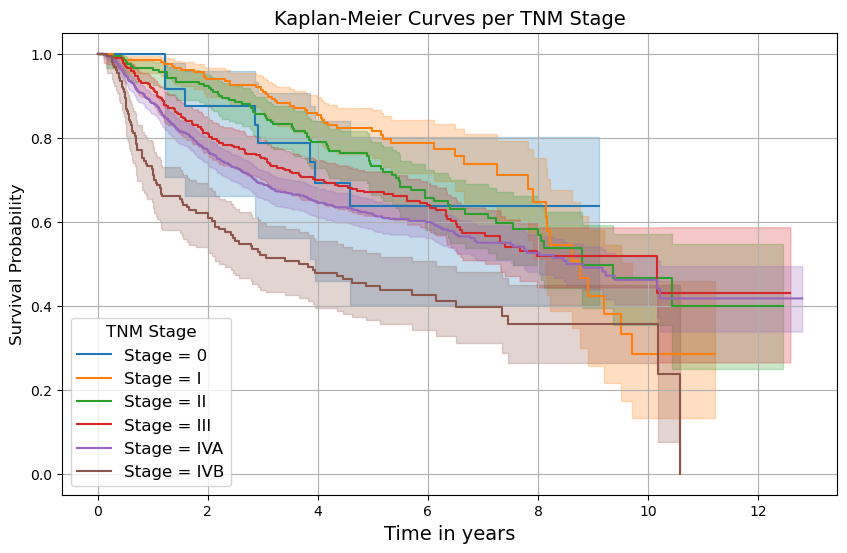}
        \label{fig:tnm_km}
    }
    \subfigure[Risk Stratification with MultiFIX]{
        \includegraphics[width=0.48\textwidth]{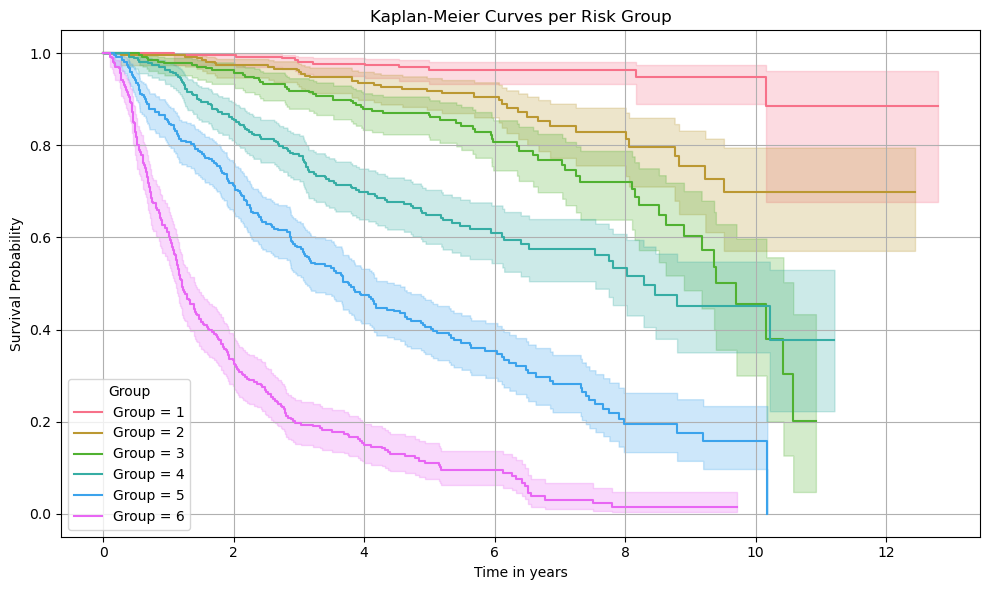}
        \label{fig:our_km}
    }
    \caption{Kaplan-Meier survival curves using two different stratification approaches: (a) TNM Staging Stratification; and (b) Risk Stratification using MultiFIX-based framework.}
    \label{fig:km_curves}
\end{figure}

The engineered features in the model, resulting from applying MultiFIX to the core data configuration, showed a high correlation with overall survival. Furthermore, these features include information that is consistent with the prognostic factors documented in the literature. The Cox regression model provides valuable insights into how each engineered feature affects the likelihood of the event (death) occurring, and it allows for personalized risk assessment for each patient. Risk stratification using risk predictions inferred from the model generated survival curves that are separable and representative of distinct groups of patients~(see Figure~\ref{fig:our_km}). This full transparency, combined with the fact that the MultiFIX survival curves are superior to those pertaining to clinically used TNM Staging, is a strong indicator of the potential for using these models in clinical practice. For the studied dataset, the TNM staging curves show poor discrimination between groups and present broad confidence intervals, especially for Stage $0$~(see Figure~\ref{fig:tnm_km}. Expert consultation suggested that TNM staging is more accurate for predicting survival endpoints directly related to cancer, such as disease-free survival. For overall survival, especially in patients with lower TNM stages, staging may be less predictive because these individuals are more likely to die from causes unrelated to cancer. This may explain the observed decline in survival rates over time for low-stage cases, as the event label for overall survival encompasses all-cause mortality rather than cancer-specific deaths. The comparison between survival curves is available in Figure~\ref{fig:km_curves}.

Expert feedback indicated that the engineered clinical features use input variables that are consistent with known prognostic factors and that patient risk stratification could provide useful complementary information alongside conventional TNM staging. The expert (CR) emphasized the potential value of easily verifiable and understandable risk stratification, not only for overall survival in head and neck cancer but also for other cancer types and endpoints, such as disease-free survival.

For the extended data configuration model, the single feature engineered from the clinical tabular data contained variables that were present in the core data configuration model. The additional clinical variables used in the extended data configuration were rarely employed in the found symbolic expressions. This can indicate that potentially relevant variables, namely the primary tumor site (only available in the extended dataset configuration), can be inferred from the image, especially from the downsampled image, since it includes more anatomical context. Feedback from the clinical expert (CR) indicated that the image activation maps could be consistent with relevant regions in the muscle tissue and soft tissues; however, the low-resolution heatmaps do not provide sufficient detail to draw definitive conclusions. The expert  (CR)further noted that, although the muscle region analyzed may capture features relevant to sarcopenia, it remains unclear whether this specific region is optimal for detecting sarcopenia, as larger muscles in the abdomen are typically assessed using more extensive CT scans. Nonetheless, the presence of sarcopenia would likely be observable even in the scanned area, should it be present.

\section{Discussion}\label{discussion}


In this study, we introduced a novel pipeline for survival analysis based on MultiFIX, an approach capable of learning multimodal interpretable models for survival prediction and risk stratification. We demonstrated our approach using data from HNC patients (specifically, the RADCURE dataset) to predict overall survival. Interpretable models learned with MultiFIX match the performance of previous studies~\cite{baseline-paper} (which do not use fully interpretable models), obtaining a C-index of $0.838~(95\% CI: 0.800,~0.869)$ for risk prediction and outperforming conventional TNM staging in risk stratification by a margin of $0.212$ with a C-index of $0.826~(95\% CI: 0.787,~0.859)$. These findings demonstrate that interpretable, clinically transparent AI models can achieve state-of-the-art performance, offering a promising pathway for integrating robust and explainable tools into oncology practice to improve patient risk assessment and decision-making. Importantly, unlike many existing methods that focus on only one or two of these aspects, MultiFIX uniquely integrates full interpretability across multiple data modalities, delivers individualized survival risk predictions, and produces interpretable stratifications into meaningful patient groups, positioning it as a comprehensive and novel contribution to the field of multimodal survival analysis.

Despite its known limitations, including the failure to account for other prognostic factors, the use of the American Joint Committee on Cancer~(AJCC) TNM staging criteria to stratify patients remains the accepted clinical guideline, in combination with other factors (co-morbidity, weight loss, ECOG PS) that position the patient at the higher or lower end of the curve (with respect to the 95\% CI). Through the use of MultiFIX, transparent patient stratification can be achieved by simultaneously incorporating clinical variables and CT imaging. The proposed approach provides an automated method to build models that can be used to identify representative patient groups by taking into account different sources of information. Moreover, these models can be explained through both the risk prediction itself as well as the engineered features for each data modality. The best-performing model we found successfully separated the patients into six distinct risk groups with statistical significance, achieving interpretability regarding how each group is defined and how each input modality contributes to the group assignment.

To the best of our knowledge, the performance analysis and interpretability of risk stratification in survival analysis using multimodal data have been overlooked in research until now. As such, this work addresses a critical gap in survival analysis that is highly relevant to clinical practice for prognosis prediction and personalized treatment. 


The novelty of the MultiFIX-based framework for survival analysis focuses on three primary objectives: engineering (non-linear) explainable features for each input data modality; using these explainable features for risk prediction with a Cox regression model; and providing a comprehensive and understandable risk stratification method. The design of our pipeline focuses on interpretability, specifically by restricting the bottleneck of engineered features per data modality, and modularity, through flexibility in changing or adding more structural blocks (for different data modalities and for fusion) to the architecture easily. In the present work, we used this modularity aspect to perform survival analysis instead of classification, as in the original paper that introduced MultiFIX~\cite{multifix-gecco}. For the extended data configuration, we added an additional feature engineering block for the downsampled version of the CT. 

Additionally to building comprehensive and powerful models, the results show minimal to no performance trade-off when replacing the DL feature engineering block for clinical variables with a symbolic expression of limited complexity, both for performance risk prediction and stratification. Moreover,  the replacement of the DL survival (fusion) block with a Cox regression model also seems to match the DL performance accurately, presenting even a consistent performance increase for both survival analysis tasks across models with engineered clinical features of different complexities. Since the DL model is optimized with a Cox Proportional Hazards loss, this is highly beneficial for the training of the models.


Expert consultation confirmed that the information used in our models is prognostic for both survival prediction and stratification. This includes the GTV regions and muscle tissues from the CT scans, as well as the majority of the clinical variables, such as age at diagnosis, smoking status, ECOG performance status, HPV status, treatment with chemoradiotherapy, T stage, N stage, and TNM stage. While it is challenging to explain the role of every single variable in detail, most of the important clinical data align well with known prognostic factors in head and neck cancer.

To broaden the clinical relevance of this work within survival analysis, our pipeline may be applied to other survival endpoints, such as disease-free survival. Additional validation on external datasets could further assess the robustness and generalizability of the developed models. However, it is important to note that the data used in this study - the RADCURE dataset - were selected primarily for illustrative purposes to demonstrate our technology, rather than as a definitive clinical tool. Questions remain regarding whether this dataset contains all relevant information, potential biases in the patient population, or missing key variables that might affect model performance. While informal expert feedback supported the clinical relevance of our approach, the model has not undergone formal prospective clinical validation. Further studies involving larger clinician panels and integration into clinical workflows are required to confirm clinical utility and generalizability.

Some limitations that can be addressed include the interpretability of image analysis, which is currently hampered by certain properties of Grad-CAM, namely its post-hoc nature, the low-resolution activation maps, and their dependence on the architecture and chosen target layer. To further enhance image interpretability, clinicians could analyze activation maps from multiple patients with different outcomes, which may help identify patterns such as the consistent involvement of certain regions in cases with poor prognosis. As future work, the image feature engineering block could benefit from an inherently interpretable architecture, providing deeper insights into which parts of the image are being used for feature construction and supporting more robust clinical understanding. Additionally, although performing well, GP-GOMEA is being used in a single-objective fashion, aiming exclusively to minimize the MSE between the GP-engineered feature and the DL feature. A multi-objective version of GP-GOMEA could evolve multiple solutions with an explicit trade-off between several objectives (such as complexity and performance). These solutions could then be shown to the clinician so that they could choose the preferred solution (e.g., which one is most understandable to them).

In conclusion, MultiFIX represents an innovative framework that bridges the critical gap between AI performance and real-world utility by providing interpretable multimodal models in domains where interpretability is of key importance. In this work, we showcased the use of a head and neck cancer dataset to demonstrate how MultiFIX can be leveraged to perform survival analysis and how the resulting models can be understood and verified by clinicians without compromising the high accuracy of black-box AI models. Importantly, the modular design of the proposed pipeline allows for easy modification of specific blocks, making it generic and ready to be used for other applications.

\section{Methods}\label{sec:methods}

This section presents the methodology used in this work to build interpretable multimodal models for survival analysis. The methods are organized into the following subsections: Subsection~\ref{subsec:radcure} describes the dataset used for this work, along with the data processing steps that were used; Subsection~\ref{subsec:survmultifix} describes the pipeline developed and the adaptations made to the original MultiFIX pipeline to perform survival analysis; Subsection~\ref{subsec:stratification} presents the part of the pipeline dedicated to risk stratification; Subsection~\ref{subsec:configs} provides the experimental configurations that were used in the pipeline to learn the models; Subsection~\ref{subsec:metrics} describes the evaluation metrics and statistical analysis performed to optimize and evaluate the learned models; Subsection~\ref{subsec:baseline} presents the baseline approaches we used to compare our models; lastly, Subsection~\ref{subsec:relevance} describes the procedure used to evaluate the clinical relevance of the learned models.

\subsection{Head and Neck Cancer Dataset}
\label{subsec:radcure}

We evaluate our approach using a publicly available dataset containing head and neck cancer patients. The RADCURE dataset~\cite{radcure-dataset} includes 3,346 HNC patients treated with radiotherapy at the University Health Network in Toronto, Canada, between 2005 and 2017. The dataset includes planning CT scans for radiotherapy and the corresponding identified GTVs. CT images were collected using systems from three different manufacturers, and standard imaging protocols were followed. Contrast enhancement was administered for some of the patients according to the clinical guidelines. Contours were manually delineated by radiation oncologists. In addition to imaging data, electronic medical records~(EMR) are linked to each patient, including demographic, clinical, and treatment information based on the 7th edition TNM staging system. A subset of this dataset was used for prognostic modeling in a collaborative challenge~\cite{radcure-challenge}, with the objective of predicting overall survival~(OS). In this subset of 2,552 patients with HNC, the inclusion criteria were treatment with radiotherapy or chemoradiation (combined radiotherapy and systemic therapy) at the Princess Margaret Cancer Centre between 2005 and 2017; the availability of planning CT imaging and target contours; at least 2 years of follow-up (or death before that period); no distant metastases at diagnosis; and no prior surgery. A train-test split was performed by the date of diagnosis: training set between 2005 and 2015 ($n=1,802$), and test set between 2016 and 2018 ($n=750$) (the dates were anonymized by a fixed offset for patient privacy). The winning solution of the challenge achieved a C-index of $0.801~(95\%~CI: 0.757-0.842)$ with deep multitask logistic regression~(MTLR) using EMR and tumor volume as a tabular feature. Further literature~\cite{baseline-paper} reported an improved C-index of $0.817~(95\%~CI: 0.766-0.846)$ using an extended version of MTLR for multimodal data, with EMR, cropped CT scans around GTV, and the respective GTV images as input. All data collection for the RADCURE dataset was approved by the Institutional Research Ethics Board, and the dataset is fully de-identified and made publicly available for research purposes. This study involved only a secondary analysis of this de-identified data and did not require additional ethical approval.

To enable both fair benchmarking against the literature and a comprehensive evaluation of our MultiFIX-based pipeline, we created two dataset configurations:
\begin{enumerate}
    \item \textbf{Core data configuration:} mimics the data selection from the multimodal MTLR solution~\cite{baseline-paper}, incorporating the following clinical variables: age, sex, ECOG performance status, smoking status, smoking frequency, T stage, N stage, TNM stage, HPV status, and chemoradiation. The CT scans and GTV masks are resampled to $128\times128\times64$ voxels, cropped using the center of mass of the primary tumor segmentation, and normalized to $[-1,1]$ using z-score normalization.
    \item \textbf{Extended data configuration:} in combination with the previously selected data, this configuration includes additional clinical variables: primary cancer site and subsite, histology type, treatment modality, total RT dose, and the number of RT fractions delivered (see Supplementary Material for complete description). Additionally, a downsampled version of the CT scan is used to provide further anatomical context, in addition to the cropped CT centered on the GTV. Both images are normalized to $[0,1]$, since this is the standard normalization protocol. This data configuration was generated after consultation with a radiation oncologist. The expert (CR) mentioned that the additional clinical variables could be of interest for survival analysis. They also emphasized that the cropped CT around the GTV could bias the model into looking exclusively into the tumor region, and that there were other regions in the medical image that could be of interest to assess the status of the patient, namely the muscle tissue to evaluate sarcopenia.
\end{enumerate}

\subsubsection{Data processing}

The CT images were resampled to $1\times1\times1~mm^3$ and cropped to $64\times128\times128$ voxels with the center of mass aligned with the GTV. For the downsampled image, resampling was adapted to $3\times3\times3~mm^3$ and cropped to $64\times128\times128$ voxels. The Hounsfield Unit~(HU) range was limited to the range $[-500,500]$ and normalized to $[-1,1]$ or $[0,1]$ according to the dataset configuration. Clinical variables were processed according to their type: numerical variables were normalized; binary and categorical variables were encoded (see Table \ S1 in Supplementary material). After processing and handling missing values, the training set included 1653 patients, with 670 patients in the test set. Overall survival was used as the endpoint for prediction. A complete description of the processing steps for the clinical variables is available in the Table \ S1 (see Supplementary Material).

\subsection{MultiFIX for Survival Analysis}
\label{subsec:survmultifix}

In this work, we employ our previously published MultiFIX framework - a Multimodal Feature engIneering approach to eXplainable AI~\cite{multifix-gecco,multifix-spie}. While MultiFIX as a framework and a concept has been published, it has only been demonstrated for classification so far and mostly on synthetic benchmarks. In this work, we adapt MultiFIX to perform survival analysis instead of classification, and we add a penalization in the loss to promote orthogonality between the engineered features. Moreover, for the first time, we use MultiFIX to analyze real-world data.

\begin{figure}
    \centering
    \includegraphics[width=0.9\linewidth]{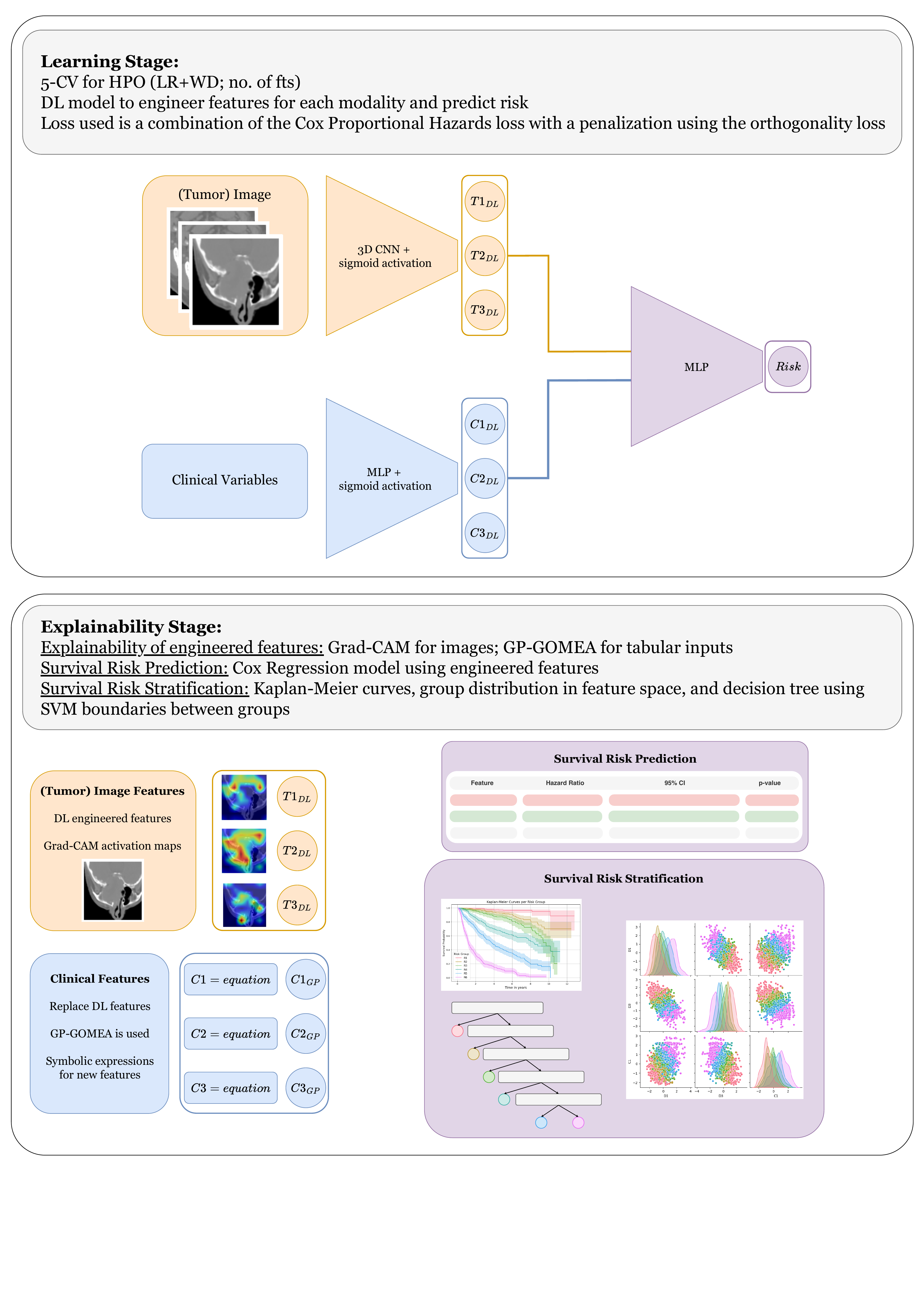}
    \vspace{-10mm}
    \caption{Pipeline overview. The learning stage (left) is used to train a DL architecture to predict the risk of event and engineer features from the available modalities. The trained model is used to extract the engineered features ($Ti_{DL}$ and $Ci_{DL}$) and use them in the explainability stage. The image features are explained using a 3D version of Grad-CAM, and the features engineered from the clinical variables are approximated using GP-GOMEA, evolving interpretable symbolic expressions ($equation$) of limited complexity. The features are used to learn a Cox regression model that can be used to make individual risk predictions and evaluate the effect of each feature for overall survival. The risk predictions are used to stratify the patients into representative risk curves, for which Kaplan-Meier survival curves are generated. The distribution of risk groups is studied in the engineered feature space to study how interactions between features correlate with the risk group assignment. Finally, a decision tree is created based on SVM boundaries to classify patients into distinct groups.}
    \label{fig:multifix}
\end{figure}

Figure~\ref{fig:multifix} illustrates an overview of the MultiFIX-based pipeline used for this work.
The generic MultiFIX pipeline consists of two stages: the learning stage and the explainability stage. In the learning stage, DL models are trained to perform a prediction task using a specific meta-architecture that is designed to automatically engineer representative and potentially relevant features from each data modality. This is achieved by using separate blocks for feature engineering per modality that output a limited number of features (up to 3 in this work) to promote simplicity and interpretability, which are then fed into a fusion block. In this work, the fusion block is a DL model capable of performing non-linear Cox regression. The entire architecture can then be trained end-to-end using common DL techniques. In the interpretability stage, each block is explained according to its specifications, and it being replaced by inherently interpretable models when possible. In our work, image features are explained post-hoc with Grad-CAM~\cite{grad-cam}; the clinical feature engineering block is fully replaced with symbolic expressions evolved with GP-GOMEA~\cite{gpg}, a model-based evolutionary algorithm for genetic programming; the fusion survival block is replaced with a Cox regression model that uses the engineered features to predict the survival risk. The resulting interpretable model provides explanations not only for the final prediction but also for each of the engineered modality-specific features. Key adaptations to the original pipeline in order to perform multimodal explainable survival analysis include: 
\begin{enumerate}
    \item The Cross Entropy Loss was replaced with a composite loss function combining the Cox proportional hazards~(CoxPH)~\cite{katzman2018deepsurv} loss with an orthogonality regularization term ($loss = CoxPH_{loss} + \lambda \times orthogonality_{loss}$), where $\lambda$ is the regularization weight parameter that enforces orthogonality constraints between the engineered features to promote feature independence and improve model interpretability.
    \item The interpretable fusion block was replaced with a Cox regression model instead of a GP-GOMEA symbolic expression because the Cox regression model is commonly used in clinical research and provides proportional hazard coefficients for each variable, which directly reflect their importance for risk prediction. To facilitate the fitting of the Cox regression model, the engineered features were normalized by subtracting the mean and linearly scaling to achieve unit variance.
    \item The adaptation of the image feature engineering block to a 3D convolutional neural network~(CNN), since the input images are now 3D CT scans, whereas only 2D images were considered in the original paper. Grad-CAM was also adapted to explain features based on 3D inputs.
    \item Hyper-parameter optimization was performed sequentially with grid-search: instead of providing a single grid to optimize learning rate~(LR), weight decay~(WD), and the number of features per modality, 5-fold cross validation~(5-CV) was used to optimize first the LR and WD, and secondly the number of features per modality, using the optimal parameters from the first step.
    \item The train-validation data splits used in 5-fold cross validation to train the models were stratified based on whether the patient died or not (if not, the sample is censored), instead of the original stratification on the target label.
    \item Hyper-parameter optimization, early stopping criteria, and best model selection was performed using the C-index as a metric, instead of the original AUROC metric.
\end{enumerate}

Figure~\ref{fig:multifix} illustrates an overview of the MultiFIX pipeline used for this work. A more detailed description of the entire workflow of the pipeline is available in the Supplementary Material, including the hyperparameters used to train the DL model, presented in Table \ S2.

\subsection{MultiFIX-based models for Patient Risk Stratification}
\label{subsec:stratification}

The MultiFIX-based pipeline is able to extract meaningful features from the available modalities and  outputs an interpretable model with which Cox-based risk prediction can be conducted by using the proportional hazard coefficients to indicate the influence of the explainable engineered features on the prediction. Despite the value of the explainable prediction scores, the stratification of patients into clinically relevant subgroups is essential for clinical experts in order to perform prognosis assessment and personalized treatment planning. 

In our work, we used the explainable engineered features and the risk predictions from the MultiFIX model to stratify patients into distinct risk groups while ensuring both separability and interpretability of the performed stratification. We implemented a quantile-based stratification approach to group patients according to their predicted risk scores. Specifically, patients are sorted by their risk predictions and divided into $N$ bins (tested for $N = \{2, 3, 4, 5, 6\}$). For each configuration, we evaluate whether the resulting survival curves are statistically different using pairwise log-rank tests with Bonferroni correction. The final number of groups is chosen as the maximum N for which all groups remain significantly distinct. The Kaplan-Meier survival curves are computed per risk group, in combination with their respective confidence intervals.

After stratifying the patients into risk groups, the group distribution is assessed in the engineered feature space, with a pairwise scatterplot that shows the relationship between each pair of features over all training patients, color-coded according to the risk group, along with a density plot of each single feature per group. To assess the separability between consecutive risk groups, we trained linear SVM classifiers at each boundary between adjacent groups. For a given boundary, patients in lower risk groups were assigned to class 0 and those in higher risk groups to class 1, transforming the multi-group problem into a series of binary classification tasks. Each SVM produced a linear decision function, which we evaluated on a held-out test set using the AUROC metric to quantify how well the groups could be distinguished. The learned linear equations (weight vector and intercept) defined the separating hyperplanes between groups, providing explicit boundary conditions. To facilitate interpretability in assigning patients to their respective groups, these SVM-derived boundaries were subsequently integrated into a decision tree, where each leaf node corresponds to a distinct risk group.

\subsection{Experimental Configurations}
\label{subsec:configs}

We trained all DL models with the Adam optimizer, a batch size of 32, and for a maximum of 100 epochs using early stopping with a patience of 15 epochs. The loss function was the composite $CoxPH_{Loss} + \lambda \cdot orthogonality$ with $\lambda = 0.001$. The DL architecture was composed of a 3D DenseNet backbone with a fully connected output layer and a sigmoid activation for the image feature engineering block, and a 3-layer MLP with batch normalization, a dropout of $0.25$, and a sigmoid activation after the output layer for the tabular feature engineering block. The survival fusion block followed the same MLP architecture without the final sigmoid activation. We use 5-CV to train the DL models and perform hyper-parameter optimization, with each train-validation data split being stratified on the ratio of censored data samples.

\begin{figure}
    \centering
    \includegraphics[width=0.8\linewidth]{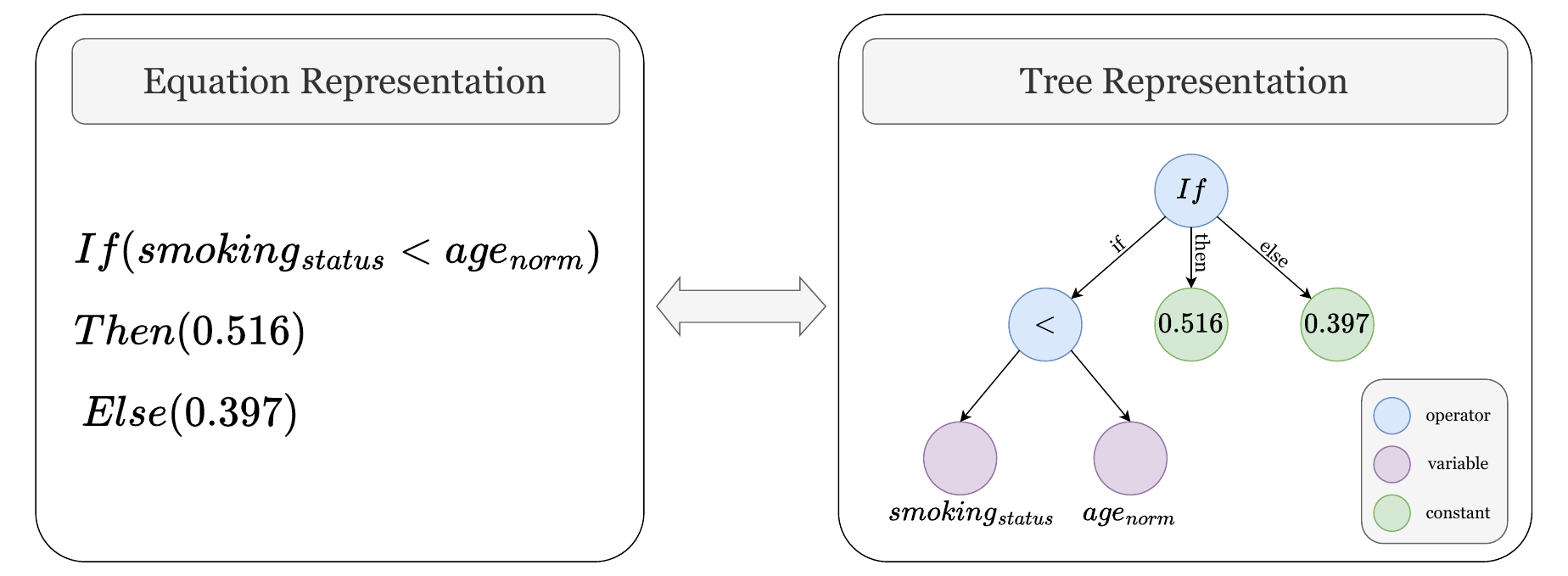}
    \caption{Overview of the same symbolic expression represented through an equation (left), and a tree (right).}
    \label{fig:gp_tree}
\end{figure}

The image-engineered features were explained by applying Grad-CAM to the last convolutional layer of the architectural block. GP represents candidate mathematical expressions as tree structures, where nodes correspond to operators, and leaves correspond to variables or constants. GP evolves these symbolic expression trees through a defined set of operators, searching for the functions that best fit the data. To approximate the DL-engineered features from the clinical variables with symbolic expressions, GP-GOMEA~\cite{gpg} was used with the interleaved multi-start scheme (IMS)~\cite{ims} for 512 generations. A mix of Boolean  and numerical  operators was used, along with the If-Then-Else operator. We tested tree template depths of 2, 3, and 4 to allow for degrees of expression complexity (Figure~\ref{fig:gp_tree} illustrates how a symbolic expression of depth 2 is represented in a tree structure). For each engineered clinical feature, experiments were run with 5 seeds per tree depth, selecting the symbolic expression with the lowest MSE. 

For the interpretable survival block, the DL architectural block is replaced by a Cox regression model using the lifelines package~\cite{lifelines} with the default settings. The engineered features are normalized between $0$ and $1$ before being used as input in the Cox regression model to ensure the same value range and, consequently, to avoid misleading proportional hazards. The features that have non-significant effects (p-value $>0.05$) are removed from the input features, and a Cox regression model is re-fitted for the final survival block. For the risk stratification of patients, quantile stratification is tested for 2, 3, 4, 5, and 6 groups, choosing the highest number of groups with statistical significance between each group using log-rank testing.

\subsection{Evaluation metrics and statistical analysis}
\label{subsec:metrics}

We optimized and evaluated the MultiFIX-based pipeline for survival analysis using several metrics: LR, WD, and the number of engineered features per modality are optimized using the average C-index over 5-fold cross-validation; both the DL and interpretable models are evaluated using the C-index and the AUC2yr on the held-out test set originally provided by the dataset; GP symbolic expressions are evaluated on the MSE in comparison to the DL-engineered clinical features; proportional hazards coefficients from the Cox regression model are statistically validated using the Wald test, where variables are considered relevant if they have coefficients with p-values $< 0.05$.

To stratify the patients into representative risk groups, statistical significance between different groups is ensured using pairwise log-rank tests with the Bonferroni correction. The performance of the stratified patient groups is achieved using the C-index and AUC2yr, utilizing the risk indices as the risk score. The AUROC is used to evaluate the performance of SVM risk group classification.

The final models are evaluated with the C-index and AUC2yr on the held-out test set using 1000 sample bootstrapping.

\subsection{Baseline Comparison}
\label{subsec:baseline}

For risk prediction, we consider 3 baselines, all of which are based on a single modality. The first one is a simple Cox regression for the clinical variables. The second is a version of the MultiFIX-based DL architecture that uses exclusively the specific feature engineering block for imaging inputs. Additionally, the published performance of the multimodal MTLR approach for single-task OS analysis~\cite{baseline-paper} is used as a multimodal baseline since the same dataset was employed, with some minor differences in patients due to the handling of missing values that could not be replicated. Since the train-test split is provided by the dataset, we ensure that the split is comparable to the baseline paper.

To stratify the patients into representative risk groups, TNM staging is used as a baseline due to its clinical acceptance and common use. To the best of our knowledge, traditional approaches often rely on binary high-low risk classifications that oversimplify the heterogeneous nature of patient populations and limit clinical utility while omitting the performance of the resulting stratification. Alternatively, survival trees can be modeled using the engineered features from our model as input, despite their limitation of defining boundaries based solely on a single feature at each split, rather than on combinations of features. We compare our group stratification performance with both the TNM staging as a clinical baseline and the use of survival trees as an academic baseline. For simplicity, we do not employ more sophisticated post-processing strategies, such as merging statistically similar groups, either in our stratification method or in the survival tree baseline.

\subsection{Clinical Relevance of the learned models}
\label{subsec:relevance}

The aim of our work is to develop a framework for multimodal survival analysis based on explainable AI, with the potential to be clinically translatable. Thus, we use the interpretability of our models to analyze whether the relevant features are correlated with known clinical findings. For clinical features, symbolic expressions can be easily used to understand which variables are selected and how they are combined, while retaining expressive power through the ability to express non-linear interactions. For imaging features, the activations can be correlated with regions of interest, such as the GTV region.

To stratify the patients into representative risk groups, the distribution of groups in the engineered feature space reveals how feature values influence each group, enabling transparent analysis and understanding of patient grouping. Comparison with the TNM staging curves can also provide insightful information.

We held two meetings with a senior radiation oncologist (CR). We presented examples of engineered features for all data modalities and survival curve outputs from several of our models to gather feedback on clinical relevance, interpretability, and potential utility in a head and neck cancer setting.


\begin{table}[h]
\begin{tabular}{c|c|c|c}
\textbf{ft} & \textbf{depth} & \textbf{best solution (out of 5 seeds)} & \textbf{MSE} \\ \hline
 &  &  &  \\
 & 2 & $If(smoking_{status} < age_{norm} )~Then(0.516)~Else(0.397)$ & 0.00580 \\
 &  &  &  \\
C1 & 3 & $\frac{4.99 + stage - chemo - HPV}{17.2}$ & 0.00357 \\
 &  &  &  \\
 & \textbf{4} & $\frac{1}{smoking_{status}+HPV+4.82} + \frac{chemo + 0.202ECOG_{PS}}{4.76 - age_{norm}}$ & \textbf{0.00193} \\
 &  &  &  \\ \hline
 &  &  &  \\
 & 2 & $If(stage < 3.33 )~Then(0.454)~Else(0.502)$ & 0.00114 \\
 &  &  &  \\
C2 & 3 & $0.013(T + stage) + 0.415$ & 0.000868 \\
 &  &  &  \\
 & \textbf{4} & $\frac{stage + 16.0}{age_{norm}+smoking_{status}+HPV+chemo+39.6}$ & \textbf{0.000447} \\
\multicolumn{1}{l|}{} & \multicolumn{1}{l|}{} &  & \multicolumn{1}{l}{}
\end{tabular}
\caption{Engineered clinical features (ft) evolved with GP-GOMEA for the core data configuration. For each of the two engineered features, the symbolic expression with the best MSE is selected for each studied complexity (maximum tree depth of 2, 3, or 4). Bold depth values and corresponding MSE indicate the symbolic expressions used in the final interpretable model.}
\label{tab:gp_expressions_baseline}
\end{table}

\begin{table}
\begin{tabular}{c|c|c|c}
\textbf{ft} & \textbf{depth} & \textbf{best solution (out of 5 seeds)} & \textbf{MSE} \\ \hline
 &  &  &  \\
 & 2 & $If(smoking_{status} < ECOG_{PS} )~Then(0.552)~Else(0.431)$ & 0.0170 \\
 &  &  &  \\
C1 & \textbf{3} & $\frac{T - chemo + age_{norm} + 3.86}{18.1 - Stage + smoking_{status}}$ & \textbf{0.0121} \\
 &  &  &  \\
 & 4 & $\frac{10.826 + ECOG_{PS} + T - chemo}{31.017 + Stage~(smoking_{status} - age_{norm})}$ & 0.00111 \\
\multicolumn{1}{l|}{} & \multicolumn{1}{l|}{} &  & \multicolumn{1}{l}{}
\end{tabular}
\caption{Clinically-engineered features evolved with GP-GOMEA for the extended data configuration. For each of the two engineered features, the symbolic expression with the best MSE is selected for each studied complexity (maximum tree depth of 2, 3, or 4). Bold depth values and corresponding MSE indicate the symbolic expressions used in the final interpretable model.}
\label{tab:gp_expressions_extended}
\end{table}

\begin{table}
\begin{tabular}{c|cc|cc}
\hline
\multirow{2}{*}{Model Description} & \multicolumn{2}{c|}{\textbf{Core Data}} & \multicolumn{2}{c}{\textbf{Extended Data}} \\ 
 & \multicolumn{1}{c|}{C-index (95\% CI)} & AUC2yr (95\% CI) & \multicolumn{1}{c|}{C-index (95/\% CI)} & AUC2yr (95\% CI) \\ \hline
\begin{tabular}[c]{@{}c@{}}DL model\\ Tumor Image Only\end{tabular} & \multicolumn{1}{c|}{\begin{tabular}[c]{@{}c@{}}0.754 \\ (0.705, 0.801)\end{tabular}} & \begin{tabular}[c]{@{}c@{}}0.775 \\ (0.720, 0.825)\end{tabular} & \multicolumn{1}{c|}{\begin{tabular}[c]{@{}c@{}}0.754 \\ (0.705, 0.801)\end{tabular}} & \begin{tabular}[c]{@{}c@{}}0.775 \\ (0.720, 0.825)\end{tabular} \\ \hline
\begin{tabular}[c]{@{}c@{}}DL model\\ Downsampled Image Only\end{tabular} & \multicolumn{1}{c|}{---} & --- & \multicolumn{1}{c|}{\begin{tabular}[c]{@{}c@{}}0.753 \\ (0.703, 0.797)
\end{tabular}} & \begin{tabular}[c]{@{}c@{}}0.777 \\ (0.720, 0.829)\end{tabular} \\ \hline
\begin{tabular}[c]{@{}c@{}}Cox\\ Clinical Variables Only\end{tabular} & \multicolumn{1}{c|}{\begin{tabular}[c]{@{}c@{}}0.788 \\ (0.744, 0.829)\end{tabular}} & \begin{tabular}[c]{@{}c@{}}0.808 \\ (0.758, 0.854)\end{tabular} & \multicolumn{1}{c|}{\begin{tabular}[c]{@{}c@{}}0.797 \\ (0.752, 0.836)\end{tabular}} & \begin{tabular}[c]{@{}c@{}}0.817 \\ (0.767, 0.862)\end{tabular} \\ \hline
MM MTLR~\cite{baseline-paper} & \multicolumn{1}{c|}{\begin{tabular}[c]{@{}c@{}}0.817 \\ (0.766, 0.865)\end{tabular}} & \begin{tabular}[c]{@{}c@{}}0.842 \\ (0.797, 0.887)\end{tabular} & \multicolumn{1}{c|}{---} & --- \\ \hline
DL model & \multicolumn{1}{c|}{\begin{tabular}[c]{@{}c@{}}0.810 \\ (0.771, 0.846)\end{tabular}} & \begin{tabular}[c]{@{}c@{}}0.827 \\ (0.783, 0.870)\end{tabular} & \multicolumn{1}{c|}{\begin{tabular}[c]{@{}c@{}}0.822 \\ (0.787, 0.856)\end{tabular}} & \begin{tabular}[c]{@{}c@{}}0.849 \\ (0.804, 0.887)\end{tabular} \\ \hline
\begin{tabular}[c]{@{}c@{}}Cox\\ DL fts\end{tabular} & \multicolumn{1}{c|}{\begin{tabular}[c]{@{}c@{}}0.837 \\ (0.801, 0.868)\end{tabular}} & \begin{tabular}[c]{@{}c@{}}0.850 \\ (0.804, 0.889)\end{tabular} & \multicolumn{1}{c|}{\begin{tabular}[c]{@{}c@{}}0.823 \\ (0.787, 0.856)\end{tabular}} & \begin{tabular}[c]{@{}c@{}}0.851 \\ (0.804, 0.887)\end{tabular} \\ \hline
\begin{tabular}[c]{@{}c@{}}Cox\\ DL fts \& Clinical Variables\end{tabular} & \multicolumn{1}{c|}{\begin{tabular}[c]{@{}c@{}}0.838 \\ (0.807, 0.870)\end{tabular}} & \begin{tabular}[c]{@{}c@{}}0.851 \\ (0.809, 0.891)\end{tabular} & \multicolumn{1}{c|}{\begin{tabular}[c]{@{}c@{}}0.818 \\ (0.778, 0.856)\end{tabular}} & \begin{tabular}[c]{@{}c@{}}0.845 \\ (0.801, 0.885)\end{tabular} \\ \hline
\begin{tabular}[c]{@{}c@{}}\textbf{MultiFIX:} Cox\\ DL img fts \& GP D2 tab fts\end{tabular} & \multicolumn{1}{c|}{\begin{tabular}[c]{@{}c@{}}0.781 \\ (0.735, 0.823)\end{tabular}} & \begin{tabular}[c]{@{}c@{}}0.802 \\ (0.746, 0.851)\end{tabular} & \multicolumn{1}{c|}{\begin{tabular}[c]{@{}c@{}}0.761 \\ (0.717, 0.802)\end{tabular}} & \begin{tabular}[c]{@{}c@{}}0.784 \\ (0.734, 0.833)\end{tabular} \\ \hline
\begin{tabular}[c]{@{}c@{}}\textbf{MultiFIX:} Cox\\ DL img fts \& GP D3 tab fts\end{tabular} & \multicolumn{1}{c|}{\begin{tabular}[c]{@{}c@{}}0.825 \\ (0.785, 0.860)\end{tabular}} & \begin{tabular}[c]{@{}c@{}}0.836 \\ (0.785, 0.879)\end{tabular} & \multicolumn{1}{c|}{\begin{tabular}[c]{@{}c@{}}\textbf{0.793} \\ \textbf{(0.753, 0.832)}\end{tabular}} & \begin{tabular}[c]{@{}c@{}}\textbf{0.817} \\ \textbf{(0.768, 0.861)}\end{tabular} \\ \hline
\begin{tabular}[c]{@{}c@{}}\textbf{MultiFIX:} Cox\\ DL img fts \& GP D4 tab fts\end{tabular} & \multicolumn{1}{c|}{\begin{tabular}[c]{@{}c@{}}\textbf{0.838} \\ \textbf{(0.800, 0.869)}\end{tabular}} & \begin{tabular}[c]{@{}c@{}}\textbf{0.854} \\ \textbf{(0.807, 0.894)}\end{tabular} & \multicolumn{1}{c|}{\begin{tabular}[c]{@{}c@{}}0.794 \\ (0.753, 0.833)\end{tabular}} & \begin{tabular}[c]{@{}c@{}}0.820 \\ (0.771, 0.865)\end{tabular} \\ \hline
\end{tabular}
\caption{Performance results for risk prediction models on held-out test set. Results from the MM MTLR are reported from their original paper without re-implementation. The model description column indicates the characteristics of each of the studied models, namely the input data (type of imaging input, or clinical variables), the model used for survival risk prediction (DL block or Cox regression~(Cox)), the engineered features used (DL-engineered features only~(DL fts), or DL image-engineered features~(DL img fts) and respective GP clinically-engineered features~(tab fts) with a specific depth~(D)). Bold models results indicate the chosen models for each data configuration. }
\label{tab:risk_performance}
\end{table}

\begin{table}
\begin{tabular}{c|ccc|ccc}
\hline
\multirow{2}{*}{} & \multicolumn{3}{c|}{\textbf{Core Data}} & \multicolumn{3}{c}{\textbf{Extended Data}} \\
 Model Description & \begin{tabular}[c]{@{}c@{}}\#\\ \end{tabular} & \begin{tabular}[c]{@{}c@{}}C-index\\ (95\% CI)\end{tabular} & \begin{tabular}[c]{@{}c@{}}AUC2yr\\ (95\% CI)\end{tabular} & \begin{tabular}[c]{@{}c@{}}\# \\ \end{tabular} & \begin{tabular}[c]{@{}c@{}}C-index \\ (95/\% CI)\end{tabular} & \begin{tabular}[c]{@{}c@{}}AUC2yr \\ (95\% CI)\end{tabular} \\ \hline
TNM Staging & 6* & \begin{tabular}[c]{@{}c@{}}0.614\\ (0.568, 0.661)\end{tabular} & \begin{tabular}[c]{@{}c@{}}0.615\\ (0.561, 0.670)\end{tabular} & 6* & \begin{tabular}[c]{@{}c@{}}0.614\\ (0.568, 0.661)\end{tabular} & \begin{tabular}[c]{@{}c@{}}0.615\\ (0.561, 0.670)\end{tabular} \\ \hline
\begin{tabular}[c]{@{}c@{}}Cox\\ Clinical Variables Only\end{tabular} & 6 & \begin{tabular}[c]{@{}c@{}}0.783 \\ (0.740, 0.822)\end{tabular} & \begin{tabular}[c]{@{}c@{}}0.805 \\ (0.755, 0.850)\end{tabular} & 6 & \begin{tabular}[c]{@{}c@{}}0.786 \\ (0.743, 0.824)\end{tabular} & \begin{tabular}[c]{@{}c@{}}0.806 \\ (0.756, 0.851)\end{tabular} \\ \hline
\begin{tabular}[c]{@{}c@{}}Cox\\ DL fts\end{tabular} & 6 & \begin{tabular}[c]{@{}c@{}}0.826 \\ (0.790, 0.859)\end{tabular} & \begin{tabular}[c]{@{}c@{}}0.841 \\ (0.794, 0.880)\end{tabular} & 6 & \begin{tabular}[c]{@{}c@{}}0.816 \\ (0.778, 0.852)\end{tabular} & \begin{tabular}[c]{@{}c@{}}0.846 \\ (0.801, 0.888)\end{tabular} \\ \hline
\begin{tabular}[c]{@{}c@{}}Cox\\ DL fts \& Clinical Information\end{tabular} & 6 & \begin{tabular}[c]{@{}c@{}}0.828 \\ (0.796, 0.859)\end{tabular} & \begin{tabular}[c]{@{}c@{}}0.842 \\ (0.801, 0.881)\end{tabular} & 6 & \begin{tabular}[c]{@{}c@{}}0.806\\ (0.765, 0.844)\end{tabular} & \begin{tabular}[c]{@{}c@{}}0.835\\ (0.789, 0.877)\end{tabular} \\ \hline
\begin{tabular}[c]{@{}c@{}}\textbf{MultiFIX:} Cox\\ DL img fts \& GP D2 tab ft\end{tabular} & 6 & \begin{tabular}[c]{@{}c@{}}0.772 \\ (0.727, 0.814)\end{tabular} & \begin{tabular}[c]{@{}c@{}}0.792 \\ (0.737, 0.841)\end{tabular} & 6 & \begin{tabular}[c]{@{}c@{}}0.749\\ (0.704, 0.788)\end{tabular} & \begin{tabular}[c]{@{}c@{}}0.770\\ (0.720, 0.819)\end{tabular} \\ \hline
\begin{tabular}[c]{@{}c@{}}\textbf{MultiFIX:} Cox\\ DL img fts \& GP D3 tab ft\end{tabular} & 6 & \begin{tabular}[c]{@{}c@{}}0.813 \\ (0.773, 0.848)\end{tabular} & \begin{tabular}[c]{@{}c@{}}0.824 \\ (0.773, 0.867)\end{tabular} & 6 & \begin{tabular}[c]{@{}c@{}}\textbf{0.787}\\ (0.750, 0.825)\end{tabular} & \begin{tabular}[c]{@{}c@{}}\textbf{0.811}\\ (0.766, 0.856)\end{tabular} \\ \hline
\begin{tabular}[c]{@{}c@{}}\textbf{MultiFIX:} Cox\\ DL img fts \& GP D4 tab ft\end{tabular} & 6 & \begin{tabular}[c]{@{}c@{}}\textbf{0.826} \\ (0.787, 0.859)\end{tabular} & \begin{tabular}[c]{@{}c@{}}\textbf{0.841} \\ (0.794, 0.882)\end{tabular} & 6 & \begin{tabular}[c]{@{}c@{}}0.7835 \\ (0.7430, 0.8225)\end{tabular} & \begin{tabular}[c]{@{}c@{}}0.8092 \\ (0.7606, 0.8549)\end{tabular} \\ \hline
\end{tabular}
\caption{Performance results for risk stratification on the held-out test set. TNM Staging is the clinical baseline to which we can compare our results, where the patient groups are stratified exclusively by the TNM Staging. For all approaches except TNM staging, the number of groups~(\#) indicates the number of significantly separable groups. The $*$ symbol in the TNM Staging results indicates that groups are not significantly separable, since the number of groups corresponds to the available stage values. Bold model results indicate the models for each data configuration.}
\label{tab:group_performance}
\end{table}

\backmatter

\newpage
\section*{Declarations}

\bmhead{Code availability}

Code developed for this study is available upon request for the purposes of editorial or peer-review assessment. Upon publication, this code will be made publicly available at a permanent online repository.

\bmhead{Supplementary Material}
Supplementary Material, including data descriptions and additional results, is presented in a separate document.

\bmhead{Funding}
This research was conducted as part of the “Uitlegbare Kunstmatige Intelligentie” project, supported by the Stichting Gieskes-Strijbis Fonds. We gratefully acknowledge the Netherlands Organization for Scientific Research (NWO) for providing a Small Compute grant (EINF-14109) on the Dutch National Supercomputer Snellius.


\bmhead{Competing Interests}
The authors declare no competing interests.

\bmhead{Author contributions}

MM, PANB, and TA conceived the study. MM performed the implementation and analysis of the presented work and wrote the manuscript. CR provided clinical expertise for the clinical relevance analysis of the learned models and revised the manuscript. PANB and TA supervised the project, provided critical feedback, and revised the manuscript. All authors read and approved the final version.



\bibliography{bibliography}

\end{document}